\documentclass[letterpaper, 11 pt, conference]{article}

\usepackage{tabularx}
\usepackage[submission]{aaai2026}
\usepackage[T1]{fontenc}

\usepackage{fancyvrb}
\usepackage{minted}
\usepackage{fvextra}
\usepackage{listings}
\fvset{breaklines=true}

\usepackage{appendix}
\usepackage{makecell}
\usepackage{graphicx} 
\usepackage{amsmath,amssymb,amsfonts}
\usepackage{textcomp}
\usepackage{multirow}
\usepackage{longtable}

\usepackage{float}
\floatstyle{plain}
\newfloat{algorithm}{H}{loa}
\floatname{algorithm}{Algorithm}

\usepackage{xcolor}
\usepackage{fancyhdr}
\usepackage{multirow}
\usepackage{booktabs}

\usepackage[ruled]{algorithm}

\usepackage[noend]{algpseudocode}

\usepackage{amsmath}
\usepackage{hhline}
\title{Learning from Prompt itself: the Hierarchical Attribution Prompt Optimization}

\author{Dongyu Chen, Jian Ma, Xianpeng Zhang, Lei Zhang, Haonan Lu, Chen Chen, Chuangchuang Wang and Kai Tang}
\date{August 2025}

\begin{document}

\maketitle

\begin{abstract}

Optimization is fundamental across numerous disciplines, typically following an iterative process of refining an initial solution to enhance performance. This principle is equally critical in prompt engineering, where designing effective prompts for large language models constitutes a complex optimization challenge. A structured optimization approach requires automated or semi-automated procedures to develop improved prompts, thereby reducing manual effort, improving performance, and yielding an interpretable process. However, current prompt optimization methods often induce prompt drift, where new prompts fix prior failures but impair performance on previously successful tasks. Additionally, generating prompts from scratch can compromise interpretability. To address these limitations, this study proposes the Hierarchical Attribution Prompt Optimization (HAPO) framework, which introduces three innovations: (1) a dynamic attribution mechanism targeting error patterns in training data and prompting history, (2) semantic-unit optimization for editing functional prompt segments, and (3) multimodal-friendly progression supporting both end-to-end LLM and LLM-MLLM workflows. Applied in contexts like single/multi-image QA (e.g., OCRV2) and complex task analysis (e.g., BBH), HAPO demonstrates enhanced optimization efficiency, outperforming comparable automated prompt optimization methods and establishing an extensible paradigm for scalable prompt engineering.
\end{abstract}
\section{Introduction}

As a result of the rapid boost of modern technology, the frequency usage of language models can be viewed as a criterion of high efficiency and convenience(\cite{huang2023languageneedaligningperception}, \cite{gong2023multimodalgptvisionlanguagemodel}, \cite{zhou2024surveyefficientinferencelarge}). However, their generalized functionality often remains inaccessible to nonexpert users, as effective interaction typically requires specialized knowledge. Prompt engineering is essential for unlocking the advanced capabilities of large language models (LLMs) for non-expert users, serving as a critical bridge between human intent and model performance. Consequently, the development of methods to automatically and efficiently optimize prompts is paramount to enhancing modern productivity. Traditional optimization, however, is grounded in continuous mathematical processes, whereas prompt refinement operates in a fundamentally discrete semantic space, necessitating tailored mechanisms.

Current automated prompt optimization strategies often address this challenge by generating new prompts or applying edits based on performance feedback. However, these methods suffer from several limitations. First, they frequently induce prompt drift, where iterative refinements fix prior failures but degrade performance on tasks the prompt previously handled successfully. In addition, generating prompts from scratch can compromise interpretability, obscuring the rationale behind the changes, and making the optimization process a black box. These issues highlight a gap in approaches that can refine prompts in a controlled, transparent, and stable manner.

To address these limitations, we propose the Hierarchical Attribution Prompt Optimization (HAPO) framework -- a novel, dynamic attribution mechanism for prompt optimization. Unlike prior approaches that statically correlate performance with benchmark scores or case-feedback, our method dynamically attributes influence based on the prompt's own semantic features and its iterative performance history. By integrating these attributes as dynamic variables within a gradient-influenced framework, and supplementing them with task-expectation grading to approximate loss, we enable a more nuanced and efficient path to better prompt design.

\textbf{Dynamic Attribution Optimization. } This paper presents a novel dynamic attribution mechanism for prompt optimization. Unlike prior approaches that statically correlate performance with benchmark scores or case-feedback, our method dynamically attributes influence based on the prompt's own semantic features and its iterative performance history. By integrating these attributes within a gradient-influenced framework and employing task-expectation grading to approximate loss, our method enables a nuanced and efficient optimization path that mitigates prompt drift.

\textbf{Semantic-unit Hierarchical Segmentation. } Also, compared to peer works, our method put more weight at the modification on the discrete semantic space, enabling targeted, interpretable edits to functional prompt segments. To follow the procedure of prompting complex tasks, we designed a process to change the prompt hierarchically, which is also a simulation of the learning rate in machine learning. We also applied the Upper Confidence Bound algorithm (UCB) \cite{han2024ucbalgorithmsmultiarmedbandits} to optimize the location and tendency of modification for long prompts. 

\textbf{Generalized Multimodal Adaptation. } In addition, with the aim of generalization in a modern environment, we attempted to deploy this mechanism into a wider range of multimodal tasks and models. We managed to apply such a strategy on tasks involving text$\leftrightarrow$image, image$\leftrightarrow$text, etc. And our method also achieves an obvious improvement in these tasks and benchmarks.

In brief, our method demonstrates compelling SoTA efficacy, achieving advanced performance in 11/12 of the evaluated scenarios while delivering a consistent average accuracy advantage of +7.21\% over the common baseline. This robust and generalized superiority across diverse reasoning and multimodal benchmarks conclusively validates it as a highly effective, model-agnostic framework for instruction optimization.

\section{Related work}

\subsection{Automatic Prompt Optimization for various tasks}
Recent prior work has developed automated methods for optimizing task-specific prompts to address the limitations of manual prompt engineering, such as APE \cite{zhou2023largelanguagemodelshumanlevel}, which generates candidates via forward/reverse LLM sampling, selects high-scoring prompts and iteratively resamples using Monte Carlo search; APO \cite{pryzant2023automaticpromptoptimizationgradient}, using textual ``gradients'' from error analysis and bandit selection for efficient refinement; and OPRO \cite{yang2024largelanguagemodelsoptimizers}, employing metaprompts to guide LLMs in generating iterative improvements. There are also evolutionary approaches such as PromptBreeder \cite{fernando2023promptbreederselfreferentialselfimprovementprompt}  and EvoPrompt \cite{guo2025evopromptconnectingllmsevolutionary}, evolving prompts via genetic algorithms; and frameworks like DSPy  \cite{khattab2023dspycompilingdeclarativelanguage}, TextGrad 
\cite{yuksekgonul2024textgradautomaticdifferentiationtext}, and  Automatic Prompt Engineering for Long Prompts \cite{guo2025evopromptconnectingllmsevolutionary}, which treat prompts as differentiable parameters for batched optimization. These methods consistently outperform manual engineering (e.g., +4–60\% on benchmarks like GSM8K and TruthfulQA) but primarily target short prompts in constrained settings, leaving complex, multi-constraint real-world applications underexplored. Evaluation typically relies on task-specific metrics (accuracy, F1) or LLM-based self-assessment. 

Compared to their approaches, our optimization process hierarchically incorporates prompt outcome scores, weakness locations in complex prompt body, and corresponding optimizing suggestion; and by plugging in the meta-prompt, this approach enables the LLM-MLLM optimizer to follow a more step-like gradient descent process, resulting in clustering of common patterns for high-quality prompts.

\subsection{Prompt quality distinguishing and refining through natural language instructions}
A recent line of research explores methodologies that leverage natural language feedback within prompts to enhance LLM performance, demonstrating effectiveness in mitigating weakness among prompting procedure. The authors of StraGo \cite{wu2024stragoharnessingstrategicguidance} joined both the good and bad cases to summarize the pro/cons through a self-trained LLM; other methodologies include task reasoning (PromptWizard\cite{agarwal2024promptwizardtaskawarepromptoptimization}), mutated word replacement (EvoPrompt \cite{guo2025evopromptconnectingllmsevolutionary}), task referencing (TAPO\cite{luo2025tapotaskreferencedadaptationprompt}), and graph optimization adapted for domain-knowledge (EGO-Prompt\cite{zhao2025autooptimizepromptsdomaintasks}). In particular, TAPO applied a task-aware evaluation strategy that connects output words towards task requirement scoring and optimization reasoning. This could lead to a more detailed attribution process, but lacks generalizability in that such word-level evaluation may only be meaningful in text tasks, rather than in thinking or other complex tasks.

In contrast to these approaches, our method implements a hierarchical framework that fully leverages the attention mechanism to consistently capture instructions. Furthermore, through iterative refinement towards prompts, our approach maintains proximity to viable candidate responses.

\subsection{MLLM's instruction-following capability}

To address deficiencies in instruction-following capabilities within MLLMs, several prior studies have developed novel methodologies. For example, some works improved this capability by implementing visual-modality token compression and cross-modality attention inhibition to mitigate image redundancy (\cite{yang2024enhancinginstructionfollowingcapabilityvisuallanguage}), while other approaches have incorporated image-based prompting skills and optimization(\cite{choi2025multimodalpromptoptimizationleverage},\cite{liu2025boostingprivatedomainunderstanding}). In addition, \cite{manas2024improvingtexttoimageconsistencyautomatic} attempted to implement specific image consistency metrics that focus only  on the instruction compliance capacity of the image generator model. 

In contrast, our method introduces a generalized strategy that circumvents the limitations of visual-information loss and narrow task specialization. This linguistically-grounded approach ensures robust performance across diverse task modules by addressing the core of the instruction-following problem, all while rigorously preserving the original structure and fidelity of the input image.

\section{Method}
\label{sec:method}

\subsection{Problem Formulation}

Let $D = \{(x_i, y_i^*)\}_{i=1}^N$ be the task dataset with $N$ samples, where $x_i$ represents input instances and $y_i^*$ represents the desired outputs.

The objective of prompt optimization is to find the improved prompt $p^*$ that minimizes the expected loss:

\begin{equation}
p^* = \arg\min_{p \in \mathcal{P}} \mathbb{E}_{(x,y^*) \sim D} \left[ \mathcal{L}(f(p,x), y^*) \right], 
\end{equation}
where:
\begin{itemize}
    \item $\mathcal{P}$ is the space of all possible prompts
    \item $f(p,x)$ is the LLM/MLLM response given prompt $p$ and input $x$
    \item $\mathcal{L}(\cdot, \cdot)$ is the loss function that measures the discrepancy between generated and desired outputs
\end{itemize}

Analogous to gradient descent in optimization, our process iteratively refines the prompt through gradient-based updates in the discrete linguistic space:

\begin{equation}
p_{t+1} = p_t - \eta \cdot \nabla_p^{\text{ling}} \mathcal{J}(p_t), 
\end{equation}
where:
\begin{itemize}
    \item $p_t$ is the prompt at iteration $t$
    \item $\eta$ is the learning rate in the linguistic space
    \item $\nabla_p^{\text{ling}}$ represents the linguistic gradient operator
    \item $\mathcal{J}(p) = \frac{1}{N} \sum_{i=1}^N \mathcal{L}(f(p,x_i), y_i^*)$ is the empirical risk
\end{itemize}

The linguistic gradient is computed through prompt analysis and response evaluation on the task-expectation grading:

\begin{equation}
\nabla_p^{\text{ling}} \mathcal{J}(p) = \mathbb{E}_{(x,y^*) \sim D} \left[ \frac{\partial \mathcal{L}}{\partial f} \cdot \frac{\partial f}{\partial p} \right], 
\end{equation}
where $\frac{\partial f}{\partial p}$ represents the functional derivative of prompt performance with respect to the LLM expectation rating, and$\frac{\partial f}{\partial p}$ represents that of the LLM response with respect to prompt variations. 

On a linguistic scale, this derivative could be explained as an attribution analysis of the impact exerted by the prompt on the LLM output. The attribution process links the model's loss to specific prompt components, which directly enables a hierarchical optimization strategy: high-attribution elements are modified first (e.g., task structure), followed by fine-grained refinements (e.g., word choice), guiding a targeted search for a better prompt. The generalized workflow is shown in Figure 1; the pseudocode is Algorithm 1.

\begin{figure*}[t]
    \centering

    \includegraphics[page=1,trim=0.6cm 5cm 0cm 0cm,clip,scale=0.8]{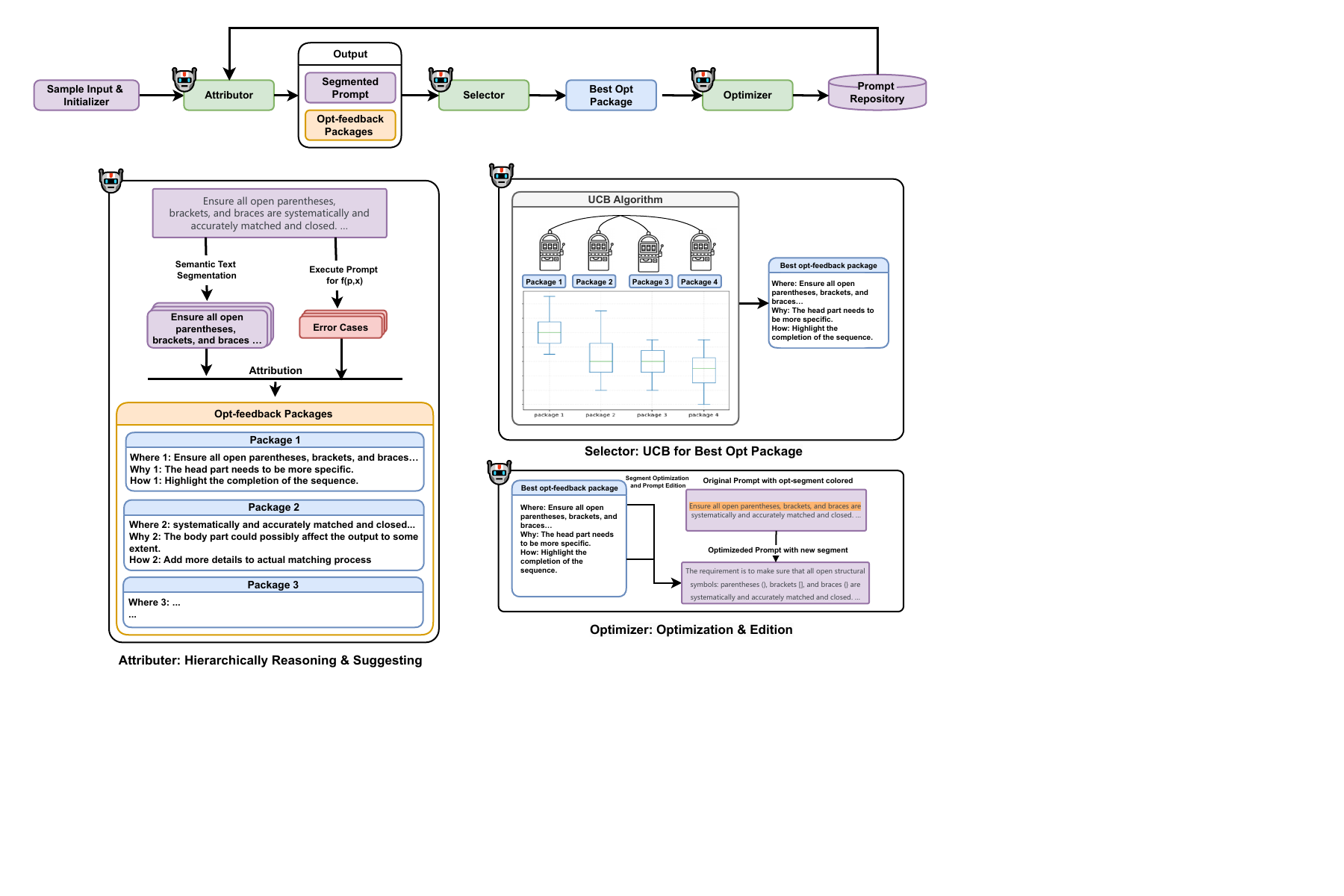}
    \caption{\textbf{Workflow of HAPO.}}
\end{figure*}

\subsubsection{Initialization Phase.}

The process begins with the initialization of a meta-prompt, into which task requirements are directly embedded. These requirements may be professionally refined by human experts, transforming preliminary rough project descriptions into formats more amenable to comprehension by specific large-scale models. This step ensures clarity and alignment with model capabilities. We also randomly selected a very small potion from the benchmark dataset as the train set $D_{train}$.

\subsection{Attributor Phase}

This phase deals with segmentation and the hierarchical attribution process of the prompt performance, generating feedback packages for the top m segments. We set $m = 4$ here.

\subsubsection{Task Result Generation. }

We will use the training-free
LLMs experimented before to produce the task results $f(p,x)$  given prompt $p$ and input $x$. To avoid irrelevant influences, the hyperparameters remain default: $temperature$: 1.0, $Top_{p}$: 1.0,
Presence Penalty and Frequency Penalty: 0.0.

\subsubsection{Semantic Text Segmentation. }
\label{sec:segmentation}
We segment $p$ into semantically coherent units using a two-stage procedure:
(i) rule-based splitting by discourse markers, section headers, list items, and delimiters; and
(ii) model-assisted refinement with a frozen instruction parser $\Pi$ that merges overly short fragments and splits run-on clauses.
Formally, $S(p)=\{u_k\}_{k=1}^K=\Pi(\mathrm{RuleSplit}(p))$.

\subsubsection{Dynamic Attribution Mechanism. }
\label{sec:attribution}
Let $\mathcal{E}_t \subset \mathcal{D}_{\mathrm{train}}$ denote the mispredicted examples in iteration $t$.
We estimate per-unit contribution scores by counterfactual occlusion with exponential smoothing:

\begin{equation}
\begin{aligned}
 s_k^{(t)} = \lambda s_k^{(t-1)} + (1-\lambda)\cdot
 \frac{1}{|\mathcal{E}_t|}\!\!\sum_{(x, y^*)\in \mathcal{E}_t}\!\!
 \Big[L\big(\mathcal{M}(x; p \setminus u_k), y^*\big) \\-  L\big(\mathcal{M}(x; p), y^*\big)\Big], 
 \label{eq:attr}
 \end{aligned}
\end{equation}
where $L$ is a surrogate loss (e.g., $0$--$1$ loss) and $p\setminus u_k$ masks $u_k$ with a neutral arm.
We augment with history-aware decay:
\begin{equation}
 \tilde{s}_k^{(t)} = \alpha_t s_k^{(t)} + (1-\alpha_t)\cdot \frac{1}{|H_k|}\sum_{(\tau,\Delta)\in H_k} \Delta \, \gamma^{t-\tau},
 \label{eq:hist}
\end{equation}
where $H_k$ stores past improvements attributed to edits on $u_k$ and $\gamma\in(0,1)$ is a temporal decay.
Top-$m$ units by $\tilde{s}_k^{(t)}$ form the actionable set.

\subsection{Selector Phase}
This phase runs a UCB process for the selection of the improved feedback package.
\subsubsection{UCB-based Edit Selection. }
\label{sec:ucb}
We model each arm as an edit candidate $a=(k,o)$ over actionable units.
Executing $a$ yields an updated prompt $p'$ and a scalar reward on a holdout dev split:
\begin{equation}
 r_t(a) = \mathrm{Acc}(p'; \mathcal{D}_{\mathrm{dev}}) - \mathrm{Acc}(p; \mathcal{D}_{\mathrm{dev}}).
\end{equation}
We maintain empirical means $\hat{\mu}_a$ and counts $n_a$.
In iteration $t$, we choose
\begin{equation}
 a_t \in \arg\max_{a} \; \hat{\mu}_a + c \sqrt{\frac{\ln t}{\max(1,n_a)}}.
\end{equation}
The procedure employs a warm-start initialization by pulling each arm once, followed by the elimination of arms whose rewards are non-positive. Given the well-separated reward distributions of the arms, a maximum iteration count $t_{max} = 100$ is sufficient.

\subsubsection{Edit Operators. }
\label{sec:operators}
We define a compact set of edit operators $\mathcal{O}$ applied to a target unit $u_k$:
(i) Replace; (ii) Insert; (iii) Delete; (iv) Reorder; (v) Refine.
An edit candidate is $a=(k,o)\in \{1,\dots,K\}\times\mathcal{O}$ that produces $p' = E_{k,o}(p)$.

\subsubsection{Multimodal Pipeline. }
\label{sec:multimodal}
For MLLM settings, the inputs include images $\{I_j\}$ and question text $x$. We extract the base64 value of the local image data as part of the MLLM request.

The optimizer constructs a joint meta-prompt that preserves the original template's core structure, populating it with the specific task requirements and constraints while applying an identical process of attribution and UCB selection. Its sole modification is an explicit annotation of the task's multimodal background at the end of the meta-prompt.

\subsection{Optimizer Phase}
This phase receives the improved opt feedback package and will split and incorporate it into the meta prompt to produce the next candidate. The meta prompt will directly include the last round's candidate, highlight the modification location in its linguistic layer, and give suggestions and reasons into structured modules. In the end, to maintain consistency, the same model will be served with this meta-prompt to generate the new candidate in the next iteration.

\subsubsection{Measuring Prompt Drift. }
\label{sec:drift}
We quantify drift as degradation on previously solved items.
Let $\mathcal{S}_{t-1}=\{i: \mathcal{M}(x_i;p_{t-1}) \text{ correct}\}$ and $\mathcal{F}_{t}=\{i\in\mathcal{S}_{t-1}: \mathcal{M}(x_i;p_t) \text{ incorrect}\}$.
We define retention and drift:
\[
 \mathrm{Retention}(t)=\frac{|\mathcal{S}_{t-1}\setminus \mathcal{F}_t|}{|\mathcal{S}_{t-1}|},\quad
 \mathrm{Drift}(t)=1-\mathrm{Retention}(t).
\]
The global drift up to $t$ is the average $\overline{\mathrm{Drift}}=\frac{1}{t}\sum_{\tau=1}^{t}\mathrm{Drift}(\tau)$.
We also trigger protective actions if $\mathrm{Drift}(t)$ exceeds a threshold for $S'$ consecutive iterations.

\subsubsection{Early Stopping and Check-pointing. }
We stop when (i) the upper-limit iteration number S (we set 20 here) is reached; (ii) no positive reward more than 0.5\% for consecutive iterations of $S'$ (we set 3 here); or (iii) the drift risk exceeds a threshold (Sect.~\ref{sec:drift}; we set it at 10\% here). We keep the improved development checkpoint and evaluate once in the test split.

\begin{algorithm}[t]
\caption{HAPO: the Hierarchical Attribution Prompt Optimization.}
\label{alg:hapo}
\begin{algorithmic}[1]
\State Initialize improved prompt $p_0$, history scores $\{s_k^{(0)}\}$, arm stats $\{\hat{\mu}_a,n_a\}$, calls $C\leftarrow 0$, $t\leftarrow 1$
\Repeat
  
  \State Run through the training set $D_{train} = \{(x_i, y_i^*)\}_{i=1}^N$ for $f(p,x_{train})$

  \State $\{u_k\}_{k=1}^K \leftarrow Seg(p_{t-1})$ \Comment{Segmentation}
  
  \State Update $\tilde{s}_k^{(t)}$ via Eqs.~\eqref{eq:attr}--\eqref{eq:hist}

  \State $\mathcal{E}_t \leftarrow$ mispredicted items on a small train slice
  
  \State Build candidate arms $\mathcal{A}_t=\{(k,o)\}$ over top-$m$ units
  \State $a_t \leftarrow \arg\max_{a\in\mathcal{A}_t} \hat{\mu}_a + c\sqrt{\tfrac{\ln t}{\max(1,n_a)}}$
  \State $p' \leftarrow E_{a_t}(p_{t-1})$; evaluate reward $r_t$ on dev
  \State Update $\hat{\mu}_{a_t}, n_{a_t}$; update improved of $\{p_{t-1},p'\}$ by dev score
  
  \State $C \leftarrow C + \text{calls used}$; $t \leftarrow t+1$
\Until{Early stopping, or iteration limit for $S$ rounds}
\State \Return improved prompt
\end{algorithmic}
\end{algorithm}

\section{Experiments}

\subsection{Implementation and Experiment setup}

\subsubsection{Models.}
Evaluation was conducted on three training-free large language models: Gemini 2.5 Pro Preview 06-05 (Gemini) \cite{comanici2025Gemini25pushingfrontier}, GPT-4o (2025-03-26) \cite{openai2024gpt4ocard}, and Qwen3-VL-Plus (2025-09-23) \cite{yang2025Qwen3technicalreport}. These models were selected for evaluation based on three principal considerations. First, their performance over time represents contemporary performance and stability, suggesting strong potential for nuanced linguistic analysis and prompt optimization. Second, each model natively supports multimodal inputs including text and images, allowing for our experiment aim. Finally, their respective APIs are engineered for high-throughput parallel computation, enabling efficient processing of large-scale benchmarks within a feasible timeframe.

\subsubsection{Benchmarks.}
Our benchmarks included BBH \cite{suzgun2022challengingbigbenchtaskschainofthought}, GSM8K \cite{cobbe2021trainingverifierssolvemath}, OCRBench V2 (OCRV2) \cite{fu2025ocrbenchv2improvedbenchmark}, and VQA2017(VQA)\cite{balanced_vqa_v2}. BBH constitutes 23 challenging text-based branches for which prior language model evaluations did not exceed average human performance; GSM8K is a dataset of 8.5K high-quality, linguistically diverse grade school math problems requiring multi-step reasoning; OCRV2 represents a large-scale bilingual text-centric benchmark of 31 diverse scenarios to evaluate visual text localization and reasoning; and VQA is a dataset containing open-ended image-to-text questions in 13 major types, demanding comprehension of vision, language, and commonsense knowledge.

BBH and GSM8K serve as representative text-to-text benchmarks, involving long-chain logical analysis and advanced mathematical problems. OCRV2 and VQA are both image-to-text benchmarks comprising high-quality visual question-answering tasks. We selected these four benchmarks to demonstrate the generalized capability of HAPO across distinct, cross-modal tasks.

From each task branch, a fixed subset of 3\% was randomly sampled. This subset was utilized throughout the optimization procedure, facilitating the computation of task accuracy at intermediate steps. These accuracy metrics provide an estimate of performance on the complete evaluation sets, thus balancing assessment cost with a reliable proxy for general capability. Upon completion of the optimization, the final instructions were evaluated on the full held-out portion of each benchmark.


\subsubsection{Baseline Methods.}

To rigorously evaluate HAPO, we compare it against six competitive baseline methods, organized into three principal categories: 

\textbf{1. Template-Based Methods.}  We include the Zero-Shot CoT prompting with prompt ``Think step by step'', and a Two-Shot CoT with two randomly selected samples in each branch  \cite{kojima2023largelanguagemodelszeroshot}. 

 \textbf{2. Auto-Generation Methods.} In this category, we compare with APE \cite{zhou2023largelanguagemodelshumanlevel} and OPRO\cite{yang2024largelanguagemodelsoptimizers}, which leverage LLMs as optimizers to automatically generate or iteratively refine prompts, though they often depend on task-specific demonstrations or meta-prompts.

\textbf{3. Gradient-Based Methods.} We encompass TextGrad \cite{yuksekgonul2024textgradautomaticdifferentiationtext} and EGO-Prompt\cite{zhao2025autooptimizepromptsdomaintasks}, which apply gradient-informed updates or evolutionary search to navigate the discrete prompt space, albeit with considerations for computational cost or sample efficiency. 

\textbf{Multimodal Adjustment. } The majority of these methods, while not originally designed, were modified to process images during prompt evaluation. However, although TextGrad can indirectly process images by integrating external MLLMs, its prompt optimization process introduces additional text-based task-loading and evaluation, inherently mismatched for multimodal tasks. Thus, TextGrad was excluded from multimodal benchmarks such as VQA and OCRV2.

\subsubsection{Evaluation.}
For evaluation, we will use an LLM grader following the benchmarks' grading rules. To avoid issues such as language patterns that affect the score performance of the grader, we chose to use a different LLM than the previous three companies for scoring; we chose Deepseek-V3 \cite{deepseekai2025deepseekv3technicalreport}; since the benchmarks all have a standard target or answer, we will fit the task, the target, and the model's output in a meta-prompt like :

\vspace{-10pt}
\begin{table}[H]
\begin{tabularx}{0.47\textwidth}{X}
\toprule
\textsl{You are a professional question-answering assessment expert. You will be given a question description (including the question itself and the answer requirements), a standard answer, and an answer; you will use this to evaluate the quality of the answer.}
\\
\\
\textsl{Question description:
\textbf{\{task\}}}

\textsl{Reference answer:
\textbf{\{target\}}}

\textsl{Answer:
\textbf{\{output\}}}
\\
\\
\textsl{Try to learn and understand the task description, and score the specific answer generated based on the task description and the reference answer to reflect whether the answer perfectly meets the question requirements in terms of steps and results, with a maximum score of 100.}\\
\bottomrule
\end{tabularx}
\end{table}
\vspace{-15pt}

An open-source, minimal, runnable prototype (including a hierarchical attributor, UCB selector, meta-hint template library, logging, and checkpoints) will be provided in our GitHub repository, along with a scaffold for reproducing experiments.

\subsection{Experiment Results}

The comprehensive experimental results are summarized in Table 1, with our method establishing an advantage, achieving an improved score in 11 out of 12 model-benchmark combinations. It delivers an average performance gain of +13.28\% over the Zero-Shot CoT baseline across all tasks and models. Specifically, it outperforms the robust OPRO optimizer by a notable margin in multimodal reasoning, such as a +2.54\% and +1.80\% percentage point advantage on the VQA benchmark (48.71\% vs. 51.25\%) and OCRV2 benchmark (54.43\% vs. 56.23\%), respectively. While TextGrad show relatively good performance in BBH benchmark, and EGO-Prompt, specialized for knowledge graph tasks through text-based expert learning, shows competitive results on BBH but underperforms on visual datasets like VQA (e.g., 41.71\% for Gemini vs. our 48.40\%, a 16\% relative improvement), our method exhibits generalized improvement. This is epitomized by its scores on GSM8K, achieving 84.81\% with Gemini (a +2.06\% lead over the second method, OPRO), 83.41\% with GPT-4o and 80.79\% with Qwen, thereby robustly validating its versatility and superior capability as a model-agnostic framework for prompt optimization.

\subsection{Model Call Analysis }

To compare the computational efficiency of the evaluated prompt optimization techniques, we conducted a comparative analysis of model calls, a primary determinant of operational cost in API-dependent environments. Beginning with the baseline, the relatively basic APE algorithm requires an average of 453.17 calls per branch. Meanwhile, OPRO exhibits a substantially higher overhead of 49,054.87 calls, a consequence of its per-sample evaluation mechanism across the entire training set. TextGrad shows significant task-dependent variance, averaging 2,365.31 calls on normal tasks, but 31,419.33 on long dataset like GSM8K. For EGO-Prompt, total calls range from approximately 3,440 for typical tasks such as BBH/VQA to 180,274 for large-scale benchmarks like GSM8K. Finally, our proposed method, HAPO, occupies an efficient position within this methodological spectrum, averaging 6.71 iterations and 2,080.10 calls per branch, achieving a relative balance between performance and resource expenditure.

\begin{table}[t]
    \centering
    \begin{tabular}{l|cccc}
        \toprule
        \midrule
        \textbf{ Method} & \textbf{BBH} & \textbf{GSM8K} & \textbf{VQA} & \textbf{OCRV2} \\
        \midrule

        \multicolumn{5}{l}{Gemini} \\
        \midrule
        \quad Zero-Shot CoT  & 70.23 & 62.45 & 39.68 & 50.06  \\
        \quad Two-Shot CoT & 71.61 & 63.81 & 41.29 & 50.48 \\
        \quad APE  & 74.18 & 64.88 & 46.24 & 58.86 \\
        \quad OPRO  & 86.95 & 82.75 & 44.10 & 59.34 \\
        \quad TextGrad  & \textbf{90.19} & 76.36 & - & - \\
        \quad EGO-Prompt  & 89.94 & 75.61 & 41.71 & 55.68 \\
        \quad Our Method  & 89.76 & \textbf{84.81} & \textbf{48.40} & \textbf{61.45} \\

        \midrule
        \addlinespace[0.3em]

        \multicolumn{5}{l}{GPT-4o} \\
        \midrule
        \quad Zero-Shot CoT  & 77.52 & 69.73 & 44.81 & 38.64 \\
        \quad Two-Shot CoT  & 77.08 & 70.17 & 43.25 & 39.01 \\
        \quad APE  & 80.63 & 72.04 & 52.06 & 44.39 \\
        \quad OPRO  & 82.86 & 78.16 & 58.94 & 47.81 \\
        \quad TextGrad  & 84.55 & 81.88 & - & - \\
        \quad EGO-Prompt  & 83.91 & 79.54 & 55.26 & 45.60 \\
        \quad Our Method  & \textbf{85.94} & \textbf{83.41} & \textbf{60.17} & \textbf{48.79} \\

        \midrule
        \addlinespace[0.3em]

        \multicolumn{5}{l}{Qwen} \\
        \midrule
        \quad Zero-Shot CoT    & 64.01 & 68.02 & 32.18 & 46.20 \\
        \quad Two-Shot CoT    & 64.46 & 69.58 & 33.63 & 45.85 \\
        \quad APE    & 61.53 & 65.07 & 32.16 & 54.18 \\
        \quad OPRO  & 73.33 & 79.35 & 43.09 & 56.15 \\
        \quad TextGrad    & 73.65 & 72.11 & - & - \\
        \quad EGO-Prompt    & 74.45 & 73.27 & 39.57 & 52.38 \\
        \quad Our Method    & \textbf{75.70} & \textbf{80.79} & \textbf{45.19} & \textbf{58.45} \\
        \bottomrule
    \end{tabular}
    \caption{\textbf{Mean Performance Across Benchmarks.} Bold numbers indicate the improved among four methods in each benchmark. Notice that TextGrad was excluded from multimodal input.}
\end{table}

\section{Prompt Optimization Case Study}


\textbf{Case 1: BBH - Date Understanding. }

This task branch requires temporal reasoning by applying a time adjustment to a given date and selecting the correct formatted output from multiple choices. The optimization path shows a language to enforce a more rigorous and error-resistant reasoning process.

In the second iteration, the prompt received a score of 95.40 and is like the following:
\begin{table}[H]
\begin{tabularx}{0.47\textwidth}{X}
\toprule
\textsl{Carefully analyze the given date information, apply the specified time adjustment, and select the calculated date in MM/DD/YYYY format with the provided choices.} \\
\bottomrule
\end{tabularx}
\end{table}
\vspace{-15pt}

In the fourth iteration with score of 96.00 it is like:
\vspace{-10pt}
\begin{table}[H]
\begin{tabularx}{0.47\textwidth}{X}
\toprule
\textsl{Carefully analyze the given date information, apply the specified time adjustment step-by-step, verify the calculated date in MM/DD/YYYY format, and select the correct option by matching it precisely with the provided choices.}\\
\bottomrule
\end{tabularx}
\end{table}
\vspace{-15pt}

In the seventh iteration, the prompt that receives a score of 99.20 is like the following:
\vspace{-10pt}
\begin{table}[H]
\begin{tabularx}{0.47\textwidth}{X}
\toprule
\textsl{Accurately interpret the given date context, perform the required time adjustment through meticulous step-by-step computation, validate the resulting date in MM/DD/YYYY format, and identify the correct answer by exact alignment with the provided options.}\\
\bottomrule
\end{tabularx}
\end{table}
\vspace{-15pt}

In this case, by incrementally enforcing a structured, multi-step computational process and demanding exact verification, the prompt guides the model to emulate a more reliable and deterministic algorithm, which is crucial for tasks requiring high numerical and logical accuracy.
\\
\\
\textbf{Case 2: OCRV2 - Text Counting. }

This task branch involves counting and outputting textual elements in an image. 

In the third iteration, the prompt, with an average grade of 28.83, is
\vspace{-10pt}
\begin{table}[H]
\begin{tabularx}{0.47\textwidth}{X}
\toprule
\textsl{Output the exact number as a numeral without any additional explanation.}\\
\bottomrule
\end{tabularx}
\end{table}
\vspace{-15pt}

In the fifth, the prompt, with average grade 43.75, is:

\vspace{-10pt}
\begin{table}[H]
\begin{tabularx}{0.47\textwidth}{X}
\toprule
\textsl{Please output the exact number without any additional explanation.}
\\
\\
\textsl{Example:}

\textsl{Question: How many times does the character `e' appear in the picture? }

\textsl{Image description: An billboard showing ``Times Square''}

\textsl{Answer: [`2', `two', `twice'] }
\\
\\
\textsl{Key note:}

\textsl{- Ensure the generated text strictly matches one of the specified target options without introducing any unlisted alternatives. }

\textsl{- Avoid introducing any additional explanations or unlisted alternatives in the output.}\\
\bottomrule
\end{tabularx}
\end{table}
\vspace{-15pt}

And in the seventh iteration the prompt, with average grade 67.89, is:

\vspace{-10pt}
\begin{table}[H]
\begin{tabularx}{0.47\textwidth}{X}
\toprule
\textsl{1. Ensure the generated text strictly matches one of the specified target options without introducing any unlisted alternatives. }

\textsl{2. Avoid redundant phrasing and maintain precision in alignment with the scoring rules.  }
\\
\\
\textsl{**Optimized Example:**  }

\textsl{Question: How many times does the character `e' appear in the picture? }

\textsl{Image description: An billboard showing ``Times Square''}

\textsl{Answer: [`2', `two', `twice']  }
\\
\\
\textsl{**Key Notes:**  }

\textsl{- The output must strictly adhere to the specified format and options.  }

\textsl{- Examples should be concise, precise, and directly aligned with the task requirements. }

\textsl{- Avoid introducing any additional explanations or unlisted alternatives in the output. }\\
\bottomrule
\end{tabularx}
\end{table}
\vspace{-10pt}
The optimization focused on output constraint and exemplar-based learning. The progression from a simple command to a detailed specification with illustrative examples and explicit guardrails (e.g., ``strictly match,'' ``avoid unlisted alternatives'') provided the model with the necessary context and constraints to align its outputs precisely with the task's evaluation criteria. The performance leap is dramatic, moving from a failing to a passing grade, achieved by evolving from a terse instruction to a richly specified prompt with demonstrations.

\section{Ablation Study}

We conducted an in-depth ablation study to evaluate the impact of various components in our proposed method. All experiments were performed using GPT-4o as the base model with default parameters unless otherwise specified. The evaluation encompasses two distinct task types: BBH's sports understanding (text-to-text reasoning) and OCRV2's reasoning VQA en (image-to-text reasoning). These datasets were selected to represent both mathematical and non-mathematical reasoning challenges while aligning with contemporary research on AI evaluation methodologies.

For evaluation metrics, we used primarily accuracy for the final output assessment to measure the stability of model performance under varying prompt conditions. 

\begin{table}[H]
\centering

\label{tab:meta_prompt_ablation}
\begin{tabular}{p{5cm}p{1cm}p{1cm}}
\toprule
Meta-Prompt Config & SU & RVE \\
\midrule
Full (Prioritizing + Reasoning) & \textbf{76.8} & \textbf{65.7} \\
$w/o.$ Prioritizing Weak Elements & 71.9 & 60.2 \\
$w/o.$ Structured Reasoning & 73.5 & 62.1 \\
$w/o.$ Both Components & 68.4 & 56.8 \\
\bottomrule
\end{tabular}
\caption{Ablation Study of Meta-Prompt Components. Here SU means sports understanding, RVE means reasoning VQA en. }
\end{table}

\subsection{The Impact of Meta-Prompt Design}

The meta-prompt's construction is critical for prompt optimization. Our default design integrates two components: Prioritizing Weak Elements (focusing on under-performance) and Structured Reasoning (explicit analysis of in-context examples). We performed an ablation study to quantify each component's contribution.

\textbf{Objective.} This part is designed to isolate the performance impact of the two core meta-prompt strategies.

\textbf{Setup. } We systematically ablated each component from the full meta-prompt and evaluated performance on BBH sports understanding and OCRV2 reasoning VQA en.

\textbf{Analysis of Outcomes. }
As shown in Table 2, the full meta-prompt achieved the highest precision. Ablating Prioritizing Weak Elements caused substantial drops (4.9\% on BBH, 5.5\% on reasoning VQA en), demonstrating its critical role. Removing Structured Reasoning led to significant but smaller reductions (3.3\% on sports understanding, 3.6\% on reasoning VQA en). Removing both components caused the most severe performance degradation (8.4\% on sports understanding, 8.9\% on reasoning VQA en), revealing a synergistic effect.

The importance of the component varied by task; Prioritizing Weak Elements was relatively more crucial for the text-based BBH task, while both components contributed more evenly to the complex visual reasoning in reasoning VQA en.

Our analytical selection strategy (utilizing the full meta-prompt) also converged faster and more stably than a randomized baseline, achieving higher final precision (76.8\% vs. 70.2\% on sports understanding; 65.7\% vs. 58.3\% on reasoning VQA en) in fewer iterations (4 vs. 8) with lower variance (±1.8\% vs. ±5.2\%), shown in Table 3.

\begin{table}[H]
\centering

\label{tab:selection_strategies}
\begin{tabular}{p{2cm}p{1.5cm}p{1.5cm}p{1.5cm}}
\toprule
Strategy & SU & RVE & $I_{Convergence}$ \\
\midrule
Analytical & \textbf{76.8} & \textbf{65.7} & \textbf{4} \\
Randomized & 70.2 & 58.3 & 8 \\
\bottomrule
\end{tabular}
\caption{Performance comparison of prompt selection strategies. Here SU means sports understanding, RVE means reasoning VQA en, and $I_{Convergence}$ means iterations
to convergence.}
\end{table}

\subsection{The Impact of Input Representation}
We also noticed that the representation of visual information requires careful design to support complex reasoning tasks, as models struggle to extract relevant information from raw visual inputs.

\textbf{Objective.} This ablation experiment is designed to assess how visual input representation forms affect OCR performance, given known AI limitations in visual reasoning.

\textbf{Setup.} We compare text-only descriptions (made by aimlessly prompting GPT-4o to caption task images, with prompt ``Briefly describe the image.'') against original images and enhanced visual representations (by adding a red box to each input image hinting the target answer), on the subset, reasoning
VQA en, of the benchmark OCRV2.

\textbf{Results. } Table 4 shows that the enhanced visual features produce the highest accuracy (65.7\%), outperforming original images (61.5\%) and text-only (58.9\%). The 6.8\% gap between text-only and original images confirms visual information is indispensable for this task, supporting findings from visual mathematical reasoning research.

\begin{table}[h]
\centering
\label{tab:input_representation}
\begin{tabular}{p{4cm}p{2cm}}
\toprule
Strategy & Accuracy (\%) \\
\midrule
Text-Only & 58.9  \\
Original Image & 61.5  \\
$w/.$ Enhanced Features & \textbf{65.7}   \\
\bottomrule
\end{tabular}
\caption{Impact of input representation on reasoning VQA en performance.}
\end{table}

\section{Reproducibility}
\label{sec:repro}
We will release anonymized code, prompts, and logs including: (i) minimal working examples per dataset, (ii) meta-prompt templates for the optimizer, (iii) configuration files ($T$, $M$), (iv) checkpoints for improved prompts and intermediate trajectories, and (v) exact preprocessing for OCR/VQA (OCR engine, box formats, resolution). All experiments use fixed seeds; we will report the versions of OS, driver, and libraries as well as the endpoints of the models. Dataset/model licenses are respected and sensitive content is filtered.

\section{Conclusion}
We introduced HAPO, a hierarchical attribution framework for prompt optimization that combines unit-level attribution, a compact edit operator set, and UCB-based selection, and extends naturally to multimodal pipelines. In fair comparisons, HAPO yields consistent gains across text and vision-language benchmarks while explicitly controlling prompt drift. We expect HAPO to serve as a practical, extensible paradigm for scalable prompt engineering and to inspire further work on attribution-driven optimization in discrete semantic spaces.
\\
\\
\textbf{Discussion.} The current HAPO method is computationally intensive, limiting its real-time use. Its evaluation also requires broader validation in professional technical domains, while its generalization to other AI models needs further study. Future work could focus on improving efficiency with adaptive prompting and early stopping, refining causal attribution methods, generalizing on domain-specific benchmarks, and expanding cross-model testing with formal drift controls.

\end{document}